\documentclass[acmtog]{acmart}
\acmSubmissionID{966}

\AtBeginDocument{%
  }

\usepackage{color, bbm}
\newcommand{\yr}[1]{\textcolor{black}{#1}}
\newcommand{\wxz}[1]{\textcolor{black}{#1}}
\newcommand{\rev}[1]{\textcolor{black}{#1}}

\setcopyright{acmlicensed}
\copyrightyear{2024}
\acmYear{2024}
\acmDOI{10.1145/3680528.3687675}

\acmConference[SA Conference Papers ’24, December 03–06, 2024, Tokyo, Japan]{SIGGRAPH Asia 2024 Conference Papers (SA Conference Papers ’24), December 03--06, 2024, Tokyo, Japan}
\acmVolume{14399}
\acmNumber{118}
\acmArticle{118}
\acmMonth{12}

\begin{document}

\title{AdR-Gaussian: Accelerating Gaussian Splatting with Adaptive Radius}

\author{Xinzhe Wang}
\authornote{Both authors contributed equally to this research.}
\orcid{0009-0006-0603-4318}
\affiliation{%
  \institution{Shanghai Jiao Tong University}
  % \city{Shanghai}
  \country{China}
}
\email{hiroxzwang@sjtu.edu.cn}

\author{Ran Yi}
\authornotemark[1]
\orcid{0000-0003-1858-3358}
\affiliation{%
  \institution{Shanghai Jiao Tong University}
  % \city{Shanghai}
  \country{China}
}
\email{ranyi@sjtu.edu.cn}

\author{Lizhuang Ma}
\authornote{Corresponding author.}
\orcid{0000-0003-1653-4341}
\affiliation{
  \institution{Dept. of Computer Science and Engineering, Shanghai Jiao Tong University}
  % \city{Shanghai}
  \country{China}
}
\email{ma-lz@cs.sjtu.edu.cn}

\renewcommand{\shortauthors}{Xinzhe Wang, Ran Yi, and Lizhuang Ma}

\begin{abstract}
3D Gaussian Splatting (3DGS) \yr{is a recent} explicit 3D representation \yr{that} has achieved high-quality reconstruction and real-time rendering of complex scenes. However, the rasterization pipeline still \yr{suffers from} unnecessary overhead resulting from avoidable serial Gaussian culling, and uneven load due to the distinct number of Gaussian \yr{to be rendered across} pixels, which \yr{hinders} wider promotion and application \yr{of 3DGS}. In order to accelerate Gaussian splatting, we propose \textit{AdR-Gaussian}, which \yr{moves part of serial culling in \textit{Render} stage into the earlier \textit{Preprocess} stage to enable parallel culling,} employ\yr{ing} adaptive radius to narrow the rendering pixel range for each Gaussian, and introduces a load balancing method to minimize thread waiting time during the pixel-parallel rendering. Our contributions are threefold, achieving a rendering speed of 310\% while maintaining \yr{equivalent} or even better quality than the state-of-the-art. Firstly, we propose \yr{to early cull Gaussian-Tile pairs of low splatting opacity based on} an adaptive radius %for projected Gaussians 
\yr{in the Gaussian-parallel \textit{Preprocess} stage}, which reduces the number of affected tile through the Gaussian bounding circle, thus reducing unnecessary overhead and achieving faster rendering speed. Secondly, we \yr{further propose early culling based on} axis-aligned bounding box for Gaussian splatting, which achieves a more significant reduction in ineffective expenses by accurately calculating the Gaussian size in the 2D directions. Thirdly, we \yr{propose} a balancing algorithm for pixel thread load, which compresses the information of heavy-load pixels to reduce thread waiting time, and enhance information of light-load pixels to \yr{hedge against rendering} quality loss. Experiments on \yr{three} datasets demonstrate that our algorithm can significantly improve the Gaussian Splatting rendering speed.
\end{abstract}





\begin{teaserfigure}
    \includegraphics[width=\textwidth]{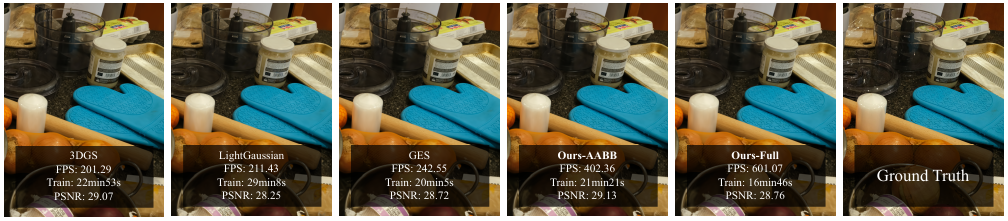}
    \caption{Our method achieves a significant improvement in rendering speed for the 3D Gaussian Splatting model with \yr{equivalent} or better quality. %while requiring similar or shorter training time compared to previous work. 
    The key to this performance is the lossless \yr{early culling of Gaussian-Tile pairs %with low splatting opacity 
    based on} adaptive radius of projected Gaussians, which narrows the rendering range for each Gaussian 
    \yr{by removing tiles with low splatting opacity.}
    %by early culling of Gaussian-Tile pairs with low splatting opacities. 
    Additionally, we propose a load balancing approach to further accelerate rendering.}
    \Description{Our full method achieves the fastest rendering speed among the existing approaches with similar quality, while our method with only adaptive radius achieves both significant improvement in rendering speed and optimal synthesis quality.}
\end{teaserfigure}

\maketitle

\section{Introduction}
Novel view synthesis (NVS) aims to \yr{generate} photorealistic rendering results \yr{of} novel views \yr{given} a set of \yr{input views}, which has \yr{attracted} significant attention due to its \yr{wide applications} in various domains, including model design \cite{DreamGaussian, GaussianEditor}, autonomous driving \cite{Matsuki:Murai:etal:CVPR2024, LightningNeRF}, and \yr{virtual reality} \cite{qian2023gaussianavatars,10204640}. 
The 3D Gaussian Splatting (3DGS) model \cite{10.1145/3592433} \yr{is a recent 3D representation that} employs %unstructured, explicit 
\yr{a set of} 3D Gaussian ellipsoids to model 3D scenes, achieving high-quality real-time rendering of complex scenes. 
However, %as a novel 3D representation method, 
\yr{the} Gaussian rasterization pipeline suffer\yr{s} from unnecessary overhead resulting from avoidable serial Gaussian culling and uneven load due to the distinct number of Gaussian to be rendered \yr{across} pixels, which limits the rendering speed of 3D Gaussian and \yr{hinders} its wider %promotion and 
applications.

%The unnecessary overhead and uneven load of the Gaussian rasterization pipeline are concentrated in the pixel-parallel \wxz{\textit{Render}} stage. The reasons are as follows: 
\yr{The Gaussian rasterization pipeline currently suffers from unnecessary overhead and uneven load due to the following reasons, which are mainly related to the \textit{Render} stage that performs pixel-parallel point-based rendering in the pipeline:}
(1) \textit{\yr{Unnecessary overhead of serial culling}}: 
Gaussian-\yr{Pixel} pairs with low splatting opacity are culled \yr{serially} during \yr{the \textit{Render}} stage, but \yr{some of the serial culling could be performed in parallel}, \yr{thus current} culling method results in a significant performance penalty for the Gaussian-serial-based alpha blending.
(2) \textit{\yr{Uneven load across pixel threads}}: 
\yr{Since the \textit{Render} stage calculates the color of each pixel in parallel, and the} number of Gaussian to be rendered \yr{serially} for each pixel \yr{are different}, \yr{different pixel} threads face distinctly different loads (computational time), %leading to variations in processing time among pixels,
which causes \yr{long thread waiting time and low rendering efficiency.} %significant waste of computing resources %based on the nature of thread waiting. 
The existing approach\yr{es \cite{fan2023lightgaussian, C3DGS, GES, girish2023eagles} that} accelerate rendering by reducing the total number of Gaussian \yr{sacrifices the rendering quality, and} fails to effectively handle the aforementioned problems, %causing a considerable deterioration in quality metrics while \wxz{yielding limited improvements in} rendering speed. 
\yr{leaving room for further improvements of rendering speed.}

To address these issues, we propose \textit{AdR-Gaussian}, which employs lossless \yr{early culling with} adaptive radius and \wxz{axis-aligned bounding box tailored for Gaussian Splatting} to narrow the rendering pixel range of each Gaussian, and \yr{proposes} a load balancing method to minimize thread waiting time.
\yr{Our} \textit{AdR-Gaussian} %accelerate\yr{s} the 3D Gaussian rendering \yr{with 
\yr{consists of two phases:} a early-culling phase and a load balancing phase. 
%pre-cull the pairs of Gaussian-\wxz{Tile} with low splatting opacity to make it streamlined from serial \wxz{\textit{Render} stage} to the parallel \wxz{\textit{Preprocess} stage}
\yr{Firstly, we propose to 
%move part of Gaussian-Tile culling that is originally conducted serially in \textit{Render} stage earlier, into the Gaussian-parallel \textit{Preprocess} stage to enable parallel culling, \textit{i.e.,} performing early culling.
move part of serial culling in \textit{Render} stage earlier into parallel culling in \textit{Preprocess} stage. We early cull Gaussian-Tile pairs with low splatting opacity during \textit{Preprocess} stage, which is conducted in parallel for each Gaussian. 
Specifically,} we propose \wxz{\textit{adaptive radius} and \textit{axis-aligned bounding box} for Gaussian Splatting} to \yr{early cull Gaussian-Tile pairs}, which \yr{achieves} accelerat\yr{ion of} Gaussian rendering without \yr{rendering quality} loss. 
Furthermore, to address the issue of uneven \wxz{computational costs} \yr{across different pixel threads} in Gaussian rendering, we \yr{propose a} \wxz{\textit{load balancing}} algorithm for pixel threads. 
%\yr{to reduce thread waiting time.}
By quantifying the variance \yr{of} the number of rendered Gaussian required for each pixel, we reduce the number of Gaussian for heavy-load pixels to reduce thread waiting time, and \yr{increase that} for light-load pixels to \yr{hedge against} the quality loss, achieving %a controlled enhancement in 
\wxz{higher rendering} efficiency.

\yr{In summary}, we make the following contributions:
\begin{itemize}
    \item We propose \textit{AdR-Gaussian}, \yr{a novel Gaussian Splatting rendering acceleration method,} which %reduces unnecessary overhead 
    \yr{moves part of serial culling in \textit{Render} stage earlier into parallel culling in \textit{Preprocess} stage}, and balances load \yr{across different} pixel thread\yr{s}. %, thus accelerating the rendering of Gaussian Splatting model.
    \item \wxz{We early cull Gaussian-Tile pairs with low splatting opacity during \yr{the Gaussian-parallel} \textit{Preprocess} stage based on adaptive radius, \yr{defined as the bounding circle radius} of \yr{an} ellipse constructed by minimum splatting opacity, \yr{which achieves acceleration without rendering quality loss}.}
    \item We \wxz{\yr{propose early culling based on} axis-aligned bounding box} tailored for Gaussian Splatting, which %\wxz{further} streamlines Gaussian culling, accelerating rendering without any loss.
    \yr{achieves different extents of culling in horizontal and vertical directions, and further improves culling efficiency.}
    \item We \yr{propose} a load balancing algorithm for pixel thread, which reduces the variance \yr{of} the number of Gaussian \yr{to be} rendered \yr{serially for} each pixel, to \yr{reduce thread waiting time and} further accelerate rendering.
\end{itemize}

\yr{Experiments on three datasets demonstrate} our approach significantly enhances the Gaussian Splatting rendering speed, achieving 310\% \yr{acceleration on average}, while \yr{achieving} equivalent or superior rendering quality. Specifically, for complex scenes in the Mip-NeRF360 dataset \cite{9878829}, our method attains an average 590FPS rendering speed on a single NVIDIA RTX3090 GPU.

\section{Related Works}
We first present a brief overview of traditional novel view synthesis approaches, followed by a discussion on the neural radiance fields (NeRF) and the 3D Gaussian Splatting (3DGS) model. Given that both of the latter two fields are extensive, we focus on the rendering acceleration \yr{methods}, which \yr{are} most pertinent to our work. For \yr{a} more comprehensive information and analysis, readers are referred to recent surveys \cite{tewari2022advances,chen2024survey}.

\subsection{Traditional Novel View Synthesis}
Early novel view synthesis methods often \yr{rely} on physical principles, such as light field sampling interpolation \cite{10.1145/237170.237199, 10.1145/237170.237200}, mesh representation \cite{10.1145/237170.237191,10.1145/383259.383309}, and voxel representation \cite{8953705,10.1145/3306346.3322980}. However, these methods \yr{are} based on discrete sampling \yr{and} suffer from issues including poor reconstruction quality, dependence on initialization data, and high spatial complexity. With the \yr{development} of neural networks, methods such as signed distance fields \cite{8954065, 9156855}, 3D occupancy fields \cite{9157495}, and differentiable rendering \cite{NEURIPS2019_b5dc4e5d} have been introduced to model 3D scenes from 2D images through implicit representation. Nevertheless, for the synthesis of high-resolution and high-quality novel \yr{views} in complex geometric scenes, all these methods proved challenging until the introduction of NeRF.

\subsection{Fast NeRF Rendering}
NeRF \cite{10.1145/3503250} employs 5D parameters including position and camera direction to obtain RGBA information %for sampling points 
via multi-layer perceptron (MLP). 
\wxz{However, due to \yr{the use of} volume rendering approach and the complexity of MLP, NeRF is inefficient, thus lots of research focus on accelerating its rendering speed.}
% Additionally, it utilizes volume rendering technology and integrates ideas from importance sampling and position encoding to enhance both efficiency and quality, resulting in high-quality 3D reconstruction results. In subsequent research, besides optimizing the neural radiation field in terms of rendering quality \cite{9878829,9710056} and training speed \cite{10.1145/3528223.3530127,9879963}, research on accelerating rendering speed is also highly valued.

% Efforts to accelerate neural radiation field rendering can be categorized into two types: reducing the number of sampling and lowering the cost per sampling operation. The purpose of NeRF's coarse-to-fine sampling approach is to minimize invalid sampling, though the weighted average sampling still falls into the invalid areas. More precisely, 
NSVF \cite{NEURIPS2020_b4b75896} introduces \yr{Octree} with empty space skipping and early ray termination strategy to minimize invalid sampling. 
RT-Octree \cite{10.1145/3610548.3618214} uses low-\yr{SPP} Monte Carlo rendering to reduce arbitrary sampling \wxz{\yr{followed by} denoising.}
% , then employs image denoising techniques.
% , achieving high-quality and highly efficient reconstruction results. 
KiloNeRF \cite{9710464} distills the large MLP into thousands of smaller MLPs, thereby reducing MLP query costs. 
Other approaches cache information using data structures \cite{9710808,9711398} or hash encoding \cite{10377265,10.1145/3592426}, further reducing query costs by minimizing MLP usage. 
Furthermore, \yr{some} work also employs GPU-friendly implementations \cite{10204779} to achieve even faster rendering speeds.

However, the\yr{se} real-time NeRF rendering \yr{often}
% are based on 
\wxz{sacrifice} other metrics, such as 
% quality degradation resulting from the reduction of arbitrary sampling, and 
slower training \yr{caused by} the establishing and updating \yr{of} data structures. 
\yr{In contrast, t}he \yr{recently proposed} 3DGS model achieve\yr{s} %comprehensive optimization of all metrics 
\yr{high-quality and real-time rendering, and fast training}
through differentiable, GPU-friendly Gaussian ellipsoids, and our method \yr{aim\rev{s} to} further improve \yr{the} rendering speed \yr{of 3DGS}.

\subsection{Fast 3DGS Rendering}
\yr{The} 3DGS model \cite{10.1145/3592433} \yr{is a recent explicit 3D representation employing a set of 3D Gaussian ellipsoids, which} %employs the explicit 3D Gaussian representation. To 
simultaneously achieve \yr{high-quality rendering}, competitive training time, and real-time rendering.
\wxz{Th\yr{e 3DGS} model trains and fits the scene via Gaussian's differentiability, \yr{and} achiev\yr{es} efficient rendering through Gaussian's explicit structure.} 
Owing to its high-quality real-time rendering capability, a \yr{large amount} of application \yr{researches have} emerged \cite{HUGS, 4DGS, SplatterImage}, while a few efforts focus on optimizing its storage \cite{C3DGS2} or synthesis quality \cite{Yu2023MipSplatting}. 

%Rendering speed, as the key contribution of 3DGS, is frequently optimized in current research through lightweight \wxz{approaches} that reduce the number of Gaussian, whose \wxz{increment} is attributed to \wxz{the} densification operations. 
\yr{Some recent researches accelerate the rendering speed of 3DGS by reducing the number of Gaussian.}
EAGLES \cite{girish2023eagles} %introduced a strategy based on the to
suppress\wxz{es} the frequency of Gaussian densification \wxz{via convergence curve}, thereby directly decelerating the growth rate of Gaussian number. 
GES \cite{GES} introduces \wxz{learning tendency from low frequency to high frequency, }
% an information learning strategy that shifts from low to high frequencies, 
limiting the creation of Gaussian for high-frequency information. 
Additionally, other methods reduce the overall number of Gaussian by pruning those with lower significance. 
LightGaussian \cite{fan2023lightgaussian} assesses Gaussian importance based on its volume, opacity, and the number of pixel affected, under which high-ratio pruning retains decent quality. 
C3DGS \cite{C3DGS} employs Gaussian scaling and opacity to generate removal mask, and incorporates masking loss into the loss function to strike a balance between synthesis quality and rendering speed.

However, %due to the lack of research on \wxz{Gaussian rasterization}, 
\yr{current fast 3DGS rendering methods lack attention to the optimization of rasterization pipeline of 3DGS, which often result in}
%rendering load 
% and the information loss caused by the reduction of Gaussian number
%current approaches yield
limited improvements in rendering speed and degenerated quality caused by the reduction of Gaussian number. 
In contrast, our \textit{AdR-Gaussian} \yr{focuses on optimizing the rasterization pipeline of 3DGS, and} can \wxz{significantly} accelerate rendering \wxz{with equivalent or better quality.}
%without compromise in quality. 

\section{Preliminaries and Analysis}

\subsection{3D Gaussian Representation}
As an %unstructured and 
explicit 3D representation, 3DGS utilizes a set of 3D Gaussian \yr{ellipsoid}s with geometric and material properties to model the scene. 

\yr{The} geometry \yr{of each} Gaussian is defined by its center position \rev{$\textbf{X}_0$}, and a 3D covariance matrix $\Sigma$ defined in world space~\cite{964490}:
\begin{equation}
    G(\textbf{\rev{X}})=e^{-\frac{1}{2}\rev{(\textbf{X}-\textbf{X}_0)}^T\Sigma^{-1}\rev{(\textbf{X}-\textbf{X}_0)}},
\end{equation}
\yr{where \rev{$\textbf{X}$ is a sampling point and} the covariance matrix is further decomposed into a scaling and a rotation matrix.}

\yr{The} material information \yr{of each Gaussian includes} opacity attribute $\sigma$, and spherical harmonics (SH) (representing color). 
To render from a specific perspective, based on the concept of Elliptic Weighted Average (EWA) \cite{1021576}, the splatting opacity $\alpha$ for each Gaussian-Pixel pair \yr{is} calculated by the opacity $\sigma$, Gaussian-pixel 2D distance $\textbf{x}$ and projected covariance matrix $\Sigma^{'}$:
\begin{equation}
    \alpha=\sigma e^{-\frac{1}{2}{\textbf{x}}^T{\Sigma^{'}}^{-1}\textbf{x}}. \label{eq:splatting-opacity}
\end{equation}

With Gaussian color $c$ calculated from SH, and splatting opacity $\alpha$, the color \yr{of each pixel is calculated} by \yr{alpha} blending $N$ related Gaussians:
\begin{equation}
    C=\sum_{i\in N}{c_i\alpha_i\prod_{j=1}^{i-1}\left(1-\alpha_j\right)}.
\end{equation}
Such an alpha blending idea is based on the neural point-based approach \cite{kopanas2021point}, and has achieved very high efficiency through a tile-based soft rasterizer.

\subsection{Gaussian Rasterization}
\label{ssec:gaussian_rasterization}

Inspired by the \yr{soft} rasterization approaches \cite{9578616}, the Gaussian rasterization \yr{adopts a tile-based and sorting renderer.
The screen is firstly splitted into $16\times 16$ tiles. And then each Gaussian is linked to the tiles it overlaps with. The Gaussians in each tile are sorted by view space depth, and then the color is calculated by alpha blending.}

\subsubsection{Rasterization Pipeline.}
\label{sssec:rasterization_pipeline}
The rasterization pipeline \yr{consists of} six \yr{stages}: \textit{Preprocess, InclusiveSum, DuplicateWithKeys, SortPairs, IdentifyTileRanges,} and \textit{Render}. Their %parallelism and 
roles are as follows:

% \begin{enumerate}
%     \item \textit{Preprocess} \yr{stage}: preprocesses each Gaussian in parallel to \yr{obtain} the \yr{perspective-related} data \yr{needed} for rendering.
%     \item \textit{InclusiveSum and DuplicateWithKeys} stages: 
% process each Gaussian in parallel. %which %aggregate the tile range information, %followed by creating an index for each tile to be rendered within each Gaussian thread.
% \yr{For each Gaussian, these stages obtain the information of which tiles it covers.}
%     \item \textit{SortPairs and IdentifyTileRanges} 
% \yr{stages}: %process each pair of Gaussian-Tile \yr{to be rendered} in parallel, %first ensuring tile and depth in order through sorting, and then determining 
% \yr{aim to sort the Gaussians in each tile by depth, and determine}
% the start and end indices of the Gaussian within each tile for rendering.
%     \item \textit{Render} stage: 
% \yr{calculates the color of} each pixel in parallel, using a point-based approach to sequentially render \yr{the} Gaussians within the \yr{corresponding} tile \yr{by} alpha blending.
% \end{enumerate}

1) \textit{Preprocess} \yr{stage}: preprocesses each Gaussian in parallel to \yr{obtain} the \yr{perspective-related} data \yr{needed} for rendering.

2-3) \textit{InclusiveSum and DuplicateWithKeys} stages: 
process each Gaussian in parallel. %which %aggregate the tile range information, %followed by creating an index for each tile to be rendered within each Gaussian thread.
\yr{For each Gaussian, these stages obtain the information of which tiles it covers.}

4-5) \textit{SortPairs and IdentifyTileRanges} 
\yr{stages}: %process each pair of Gaussian-Tile \yr{to be rendered} in parallel, %first ensuring tile and depth in order through sorting, and then determining 
\yr{aim to sort the Gaussians in each tile by depth, and determine}
the start and end indices of the Gaussian within each tile for rendering.

6) \textit{Render} stage: 
\yr{calculates the color of} each pixel in parallel, using a point-based approach to sequentially render \yr{the} Gaussians within the \yr{corresponding} tile \yr{by} alpha blending.

\begin{figure*}
  \centering
  \includegraphics[width=0.9\linewidth]{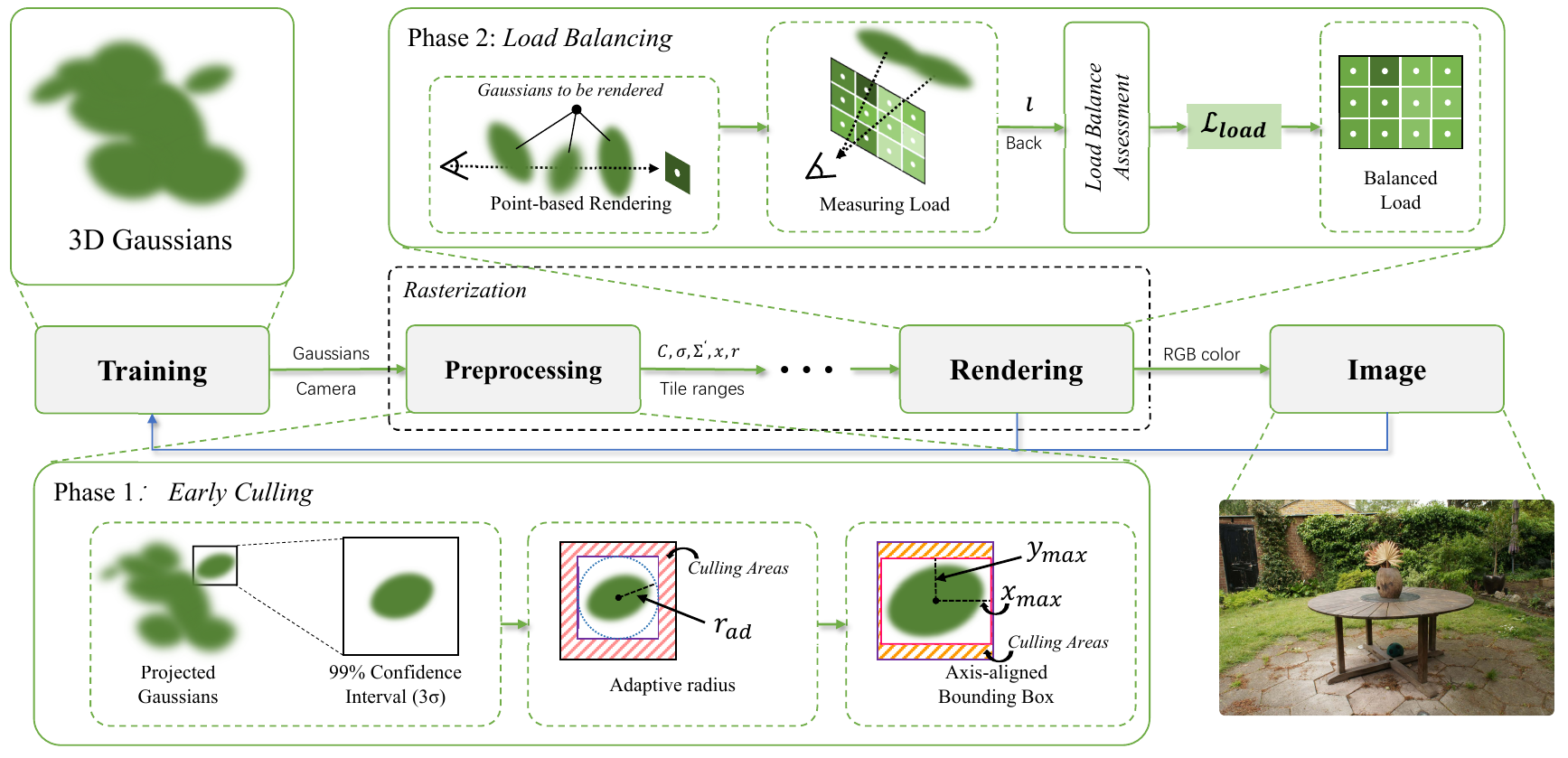}
  \caption{Pipeline overview. We \yr{propose} \textit{AdR-Gaussian}, designed to accelerate the rendering process of 3DGS. In this approach, we propose to \yr{move part of serial culling in \textit{Render} stage earlier into parallel culling in \textit{Preprocess} stage. We} \wxz{early} cull Gaussian-Tile with low splatting opacity \yr{based on} \wxz{adaptive radius}, \yr{which is} the bounding circle radius of the ellipse constructed by minimum splatting opacity. 
  Furthermore, we employ \textit{axis-aligned bounding box} tailored for Gaussian Splatting to improve culling efficiency.
  %, making it streamlined from serial \textit{Render} \yr{stage} to the parallel \textit{Preprocess} \yr{stage}. 
  \wxz{To address the uneven load issue}, we \yr{propose a} load balancing algorithm for pixel threads, which reduces the number of Gaussian for heavy-load pixels to reduce thread waiting time, and \yr{increases that} for light-load pixels to \yr{hedge against} quality loss.}
  \Description{After Gaussians are transformed from training stage to raster, we first use adaptive radius \wxz{and axis-aligned bounding box} to early-cull those pairs of Gaussian-Tile with low splatting opacity during the \textit{Preprocess} stage \wxz{, which streamline part of the culling, resulting in lossless acceleration of rendering speed. When these Gaussians are rendered, we count the load of each pixel thread and then return it to the training stage to control it through loss function, which reduces the number of Gaussian to be rendered for heavy-load pixels and increases that for light-load pixels.}}
  \label{fig:pipeline}
\end{figure*}

\subsubsection{Rasterization Costs Analysis.}
\label{sssec:rasterization_cost_analysis}
The computational cost $\mathcal{E}$ required in the Gaussian rasterization pipeline can be categorized into three parts, depending on the degree of GPU parallelism:
\begin{equation}
    \mathcal{E} = {\mathcal{E}}_g + {\mathcal{E}}_n + {\mathcal{E}}_p,
    \label{eq:rasterization_cost}
\end{equation}
where (1) ${\mathcal{E}}_g$ denotes the Gaussian-level parallel cost \yr{of the 1-3th stages}, for preprocessing Gaussians to obtain their color and other perspective related attributes; 
(2) ${\mathcal{E}}_n$ represents the Gaussian-Tile pair parallel cost \yr{of the 4-5th stages}, for identifying %tile range for each Gaussian
\yr{Gaussian range for each tile}; 
and (3) ${\mathcal{E}}_p$ represents the pixel-level parallel cost \yr{of the 6-th stage}, for the point-based rendering.

Due to the serial alpha blending of Gaussians in the \textit{Render} stage, and the complexity of calculating the splatting color for each Gaussian-Pixel pair, this stage has the largest \yr{computational time} proportion\yr{, \textit{i.e.,} ${\mathcal{E}}_p$ takes the largest proportion (over 50\%) in $\mathcal{E}$}. 
To accelerate \yr{the \textit{Render}} stage, we propose \textit{AdR-Gaussian}, which culls pairs of Gaussian-Tile with low splatting opacity \yr{earlier in \textit{Preprocess} stage,} and balances the computational costs of pixel thread.

\section{Method}

\subsection{Overview}

%Our main goal is to reduce the computational cost \yr{of} the Gaussian rasterization pipeline. %\yr{(Sec.~\ref{ssec:gaussian_rasterization})}, which can be categorized into three parts depending on the degree of GPU parallelism:
%\begin{equation}
%    \mathcal{E} = {\mathcal{E}}_g + {\mathcal{E}}_n + {\mathcal{E}}_p,
%\end{equation}
%where ${\mathcal{E}}_g$ denotes the Gaussian-level parallel cost for preprocessing Gaussians to obtain their color and other perspective related attributes, 
%${\mathcal{E}}_n$ represents the Gaussian-Tile pair parallel cost for identifying %tile range for each Gaussian
%\yr{Gaussian range for each tile}, and 
%${\mathcal{E}}_p$ represents the pixel-level parallel cost for the point-based rendering.

Our main goal is to reduce the computational cost $\mathcal{E}$ \yr{of} the Gaussian rasterization pipeline \yr{(Sec.~\ref{ssec:gaussian_rasterization})}, \yr{which can be decomposed into three types of costs as in Eq.~\eqref{eq:rasterization_cost}.} 
Among \yr{these} three types of cost\yr{s}, the \yr{pixel-level parallel cost ${\mathcal{E}}_p$ of the \textit{Render} stage (Sec.~\ref{sssec:rasterization_pipeline}) takes the largest proportion. 
This is because the} pixel-parallel \yr{\textit{Render} stage has} a high degree of seriality in the handling of Gaussians, 
\yr{\textit{i.e.,} to calculate pixel color based on} alpha blending, it necessitates the sequential traversal of all Gaussians within the corresponding tile. 
%thereby resulting in a significant overhead for this aspect.
\yr{Therefore, we aim to accelerate the pixel-parallel \textit{Render} stage, by moving part of serial culling earlier into Gaussian-parallel stages, and balancing the computational costs across parallel pixel threads.}

%While most existing methods of faster rendering aim to reduce the overall number of Gaussian, which reduces the number of serial processed Gaussian in pixel parallel module, our method focuses on early culling and load balancing to achieve higher parallelism and maintain or even improve synthesis quality.

\yr{As shown} in Fig.~\ref{fig:pipeline}, the \yr{pipeline} of our \textit{AdR-Gaussian} \yr{consists of} two distinct phases:

% \subsubsection{Phase 1: Early Culling (Sec.~\ref{ssec:adaptive_radius},~\ref{ssec:aabb}).}
\textbf{Phase 1: Early Culling (Sec.~\ref{ssec:adaptive_radius},~\ref{ssec:aabb}).} 
%We \yr{move part of Gaussian-Tile culling that is originally conducted serially in \textit{Render} stage earlier, into the \textit{Preprocess} stage to enable parallel \yr{culling}. \yr{In this way, we convert serial culling into parallel culling and conduct culling earlier, reducing the number of Gaussians to be processed in the later stages}, and increasing rendering efficiency.
%This process is streamlined from the serial rendering module to the parallel preprocessing module, enhancing computational parallelism and leading to efficient computation.
\yr{We move part of serial culling in \textit{Render} stage earlier into parallel culling in \textit{Preprocess} stage.
We early cull Gaussian-Tile pairs with low splatting opacity during \textit{Preprocess} stage, which is conducted in parallel for each Gaussian, and reduces the number of Gaussian-Tile to be processed in the later stages.
Specifically, we propose two early culling algorithms: 1) we early cull} pairs of Gaussian-Tile with \yr{low splatting} opacity \yr{based on adaptive radius, which is the radius of the} bounding circle \yr{of an ellipse constructed by minimum splatting opacity.} %for projected Gaussians.  
 2) Furthermore, we employ axis-aligned bounding box to improve culling efficiency and achieve higher rendering speed.

% \subsubsection{Phase 2: Load balancing (Sec.~\ref{ssec:load_balancing}).}
\textbf{Phase 2: Load balancing (Sec.~\ref{ssec:load_balancing}).} 
\yr{For \textit{Render} stage, since the color of each pixel is calculated in parallel, and the computational costs of different pixel threads are uneven, the overall efficiency largely depends on heavy-load pixel threads.} 
We \yr{propose to balance} load \yr{across} pixel threads, reducing the number of rendered Gaussian for heavy-load pixels to \yr{minimize} thread waiting, and increasing that for light-load pixels to hedge against quality loss, \yr{which achieves} more efficient rendering with \yr{equivalent} or better quality.

% \wxz{With \yr{the} early culling and load balancing \yr{designs}, 
% % both phases, 
% \yr{our \textit{AdR-Gaussian}} can achieve \yr{a 310\%} higher rendering efficiency \yr{on average}.}

\subsection{
\yr{Early Culling with Adaptive Radius}}
%Gaussian Bounding Circle With Adaptive Radius}
\label{ssec:adaptive_radius}

\yr{Since the Gaussians in each tile are rendered in serial,} \wxz{the} %in
\yr{\textit{Render} stage} takes the largest load proportion, %due to the serial rendering of Gaussians. 
\yr{which motivates us to accelerate this stage.}
\yr{We observe that the cull of Gaussians (removing Gaussians of low splatting opacity) are conducted in \textit{Render} stage: }
%Based on the splatting algorithm~\cite{10.1145/383259.383300}, the splatting opacity of a 2D ellipse is comprised of its intrinsic opacity $\sigma$ and a splatting coefficient \yr{(Eq.~\eqref{eq:splatting-opacity})}, The splatting coefficient is calculated based on the projected covariance $\Sigma^{'}$ and Gaussian-Pixel 2D distance $\textbf{x}$\yr{, which is pixel-related}, % Because
\wxz{The }
\yr{splatting opacity of each Gaussian is calculated by Eq.~\eqref{eq:splatting-opacity}, based on Gaussian-Pixel 2D distance $\textbf{x}$, which is pixel-related,} 
\wxz{ and pixel information can only be obtained from \textit{Render} stage.}
% the calculation of this splatting opacity is carried out during the \textit{Render} stage.
%However, \yr{in the current pipeline, the cull of Gaussians ,} some Gaussians \yr{with low splatting opacity} will be removed and contributed nothing, %due to low splatting opacity, 
%which denotes that these Gaussians can be culled earlier to avoid serial processing, such serial method can result in unnecessary overhead.
\yr{However, since the process of each Gaussian in \textit{Render} is serial, which makes the culling also serial. We regard the culling process can be parallel, and propose to move part of the culling earlier, into previous \textit{Preprocess} stage to avoid serial culling and enable parallel culling.}

\begin{figure}[t]
  \centering
  \includegraphics[width=0.8\linewidth]{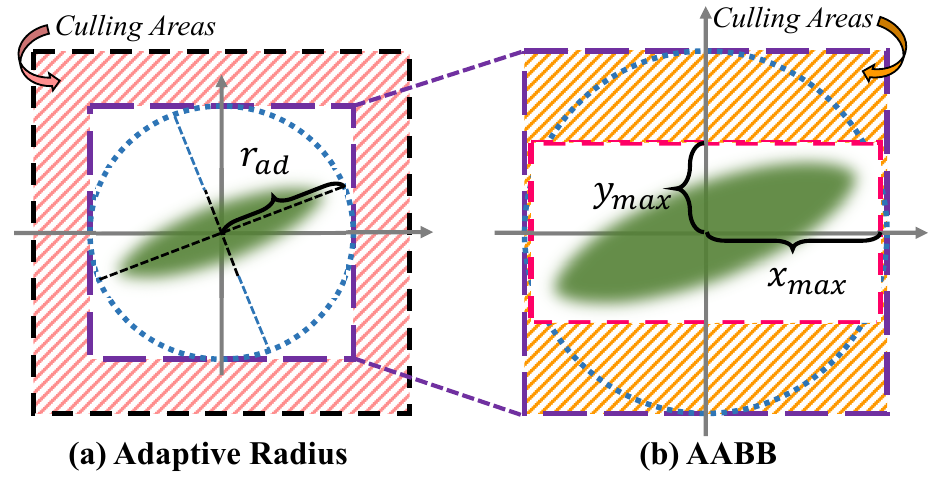}
  \caption{The rendering range of early culling approaches: \yr{(a) Adaptive Radius, (b) AABB for Gaussian}. 
  The original rendering range, \yr{\textit{i.e.,} the bounding box of 99\% confidence interval, is shown as the black square, which will be further culled in \textit{Render} stage based on splatting opacity.} %might be considerable (in black), with redundant areas that will not be rendered (in red). 
  \yr{W}e propose \yr{to early cull in \textit{Preprocess} stage with (1)} adaptive radius, \yr{which is the bounding circle (blue) radius of an ellipse constructed by minimum splatting opacity, and remove the redundant red areas}, resulting in a smaller rendering range (in purple); \yr{and (2) we} further \yr{propose to cull} with axis-aligned bounding box (in rose red) for Gaussian Splatting, \yr{which} achieves different extents of culling in 2D directions, significant\yr{ly} accelerat\yr{ing} rendering speed.}
  \Description{Different rendering extents of a projected Gaussian. The original rendering range \yr{is the bounding box of 99\% confidence interval, which} might be quite large. \yr{We propose to early cull with adaptive radius and AABB for Gaussian. } \wxz{The adaptive radius is calculated from an ellipse constructed by minimum splatting opacity so that we can remove the pairs of Gaussian-Tile with low opacity earlier. To further accelerate rendering, we achieve different extents of culling in 2D directions by proposing the axis-aligned bounding box tailored for Gaussian Splatting. Both of these approaches streamline the culling of Gaussian from Gaussian-serial \textit{Render} stage to Gaussian-parallel \textit{Preprocess} stage, which leaves a smaller pixel range for each Gaussian without loss, thus achieving higher rendering speed.} 
  % while that of the bounding circle and axis-aligned bounding box can be smaller. Our rendering extent is based on the circumscribed square of the bounding circle, whose rendering range is still quite small.
  }
  \label{fig:bboxs}
\end{figure}

Our goal is to early cull pairs of Gaussian-\yr{Tile} with extremely low splatting opacity \yr{in parallel in \textit{Preprocess} stage,} \yr{to} avoid unnecessary \yr{serial processing} and achieve faster rendering speed. 
To \yr{achieve} this, we propose adaptive radius for projected Gaussians, which computes the bounding circle \yr{for each Gaussian,} and early culls \yr{tiles} out of bounding circle during the \textit{Preprocess} stage, where Gaussians are processed in parallel. 
Compared to the original rendering range based on $99\%$ confidence interval \yr{(black square)}, the rendering range \yr{of} the bounding circle is shown 
\wxz{as the \yr{purple} square}
in Fig.~\ref{fig:bboxs}a. 
Owing to the constraints of the rasterization pipeline, the actual rendering extent is the circumscribed square of the bounding circle.

To get the bounding circle, we compute projected Gaussian's adaptive radius based on the splatting opacity, instead of the original \yr{$99\%$ confidence interval} (calculated from the standard deviation of a 2D Gaussian distribution). 
\yr{Although the splatting opacity $\alpha$ is yet unknown during the \textit{Preprocess} stage, we know a lower bound of splatting opacity, which is a %hyperparameter 
predefined minimum splatting opacity constant $\alpha_{low}$
set in the original 3DGS model:}
%That is, the calculation of the adaptive radius starts with the requirement for a minimum level of splatting opacity, which is a hyperparameter set in the original 3D Gaussian Splatting model:
% \begin{equation}
%     \alpha_i\geq\alpha_{low},
%     \label{eq:alpha_low}
% \end{equation}
\rev{$\alpha_i\geq\alpha_{low}$,}
where ${\alpha}_{i}$ \yr{is} the splatting opacity of Gaussian $i$ at the current pixel. %, and ${\alpha}_{low}$ is a predefined minimum splatting opacity constant.
%According to Eq.~\eqref{eq:splatting-opacity}, the splatting opacity ${\alpha}_{i}$ is related to Gaussian opacity $\sigma$, projection covariance ${\Sigma}^{'}$, and 2D distance $\textbf{x}$ between Gaussian and Pixel, thus 
\yr{By substituting Eq.~\eqref{eq:splatting-opacity} into} \rev{this} 
% the above Eq.~\eqref{eq:alpha_low} 
\yr{inequality, we get the following expression:}
%the aforementioned inequality can be expressed in the following manner:
\begin{equation}
    {\textbf{x}}^T{\Sigma^{'}}^{-1}\textbf{x}\le2  ln\left(\frac{\sigma_i}{\alpha_{low}}\right).
    \label{eq:ieq-fst}
\end{equation}

Since 2D distance $\textbf{x}$ can be \yr{expressed} as its 1D components $\left(x, y\right)$, \yr{and} \wxz{covariance} \yr{$\Sigma^{'}$ can be expressed as $[\Sigma^{'}_X, \Sigma^{'}_{XY}; \Sigma^{'}_{XY}, \Sigma^{'}_Y]$,} inequality \rev{Eq.~\eqref{eq:ieq-fst}} represents an ellipse, wherein \yr{$x,y$} serve as variables:
\begin{equation}
    A x^2+B y^2+C x y+D\le0, \label{eq:2d-ellipse}
\end{equation}
where
\begin{displaymath}
    A={\Sigma^{'}_Y}, \ B={\Sigma^{'}_X}, \ C=-2 {\Sigma^{'}_{XY}},
\end{displaymath}
\begin{displaymath}
    D=-2  \left({\Sigma^{'}_X}{\Sigma^{'}_Y}-{\Sigma^{'}_{XY}}^2\right)  ln\left(\frac{\sigma_i}{\alpha_{low}}\right).
\end{displaymath}

\yr{We then define the adaptive radius of a projected Gaussian as the bounding circle radius of this ellipse, \textit{i.e.,} half of the ellipse's major axis length.}
%For a projected Gaussian, its adaptive radius should depend on the maximum Euclidean distance from the pixel that satisfies the above inequality, that is, the adaptive radius is precisely equal to the length of the semi-major axis of the above ellipse:
%\begin{equation}
%    r_{ad}={\sqrt{{x}^2+{y^2}}}_{max}.
%\end{equation}
% To calculate the semi-major axis of the ellipse, we first compute its covariance ${\Sigma}_{ad}$ and eigenvalue ${\lambda}_{ad}$ according to the characteristic equation$\left|{\Sigma}_{ad}-{\lambda}_{ad} I\right|=0$:
% \begin{equation}
%     {\Sigma}_{ad}=\left(\begin{matrix}-A/D&-C/2D\\-C/2D&-B/D\\\end{matrix}\right),
% \end{equation}
% \begin{equation}
%     {\lambda}_{ad}=\frac{-\left(A+B\right)\pm\sqrt{\left(A-B\right)^2+C^2}}{2D}.
% \end{equation}
% Subsequently, according to the standard equation of ellipse, the adaptive radius is equal to the arithmetic square root of the reciprocal of the smaller eigenvalue.
% \begin{equation}
%     {\lambda}^{'}_{max} = \left(\frac{{\Sigma^{'}_X}+{\Sigma^{'}_Y}}{2}+\sqrt{\left(\frac{{\Sigma^{'}_X}+{\Sigma^{'}_Y}}{2}\right)^2-\left({\Sigma^{'}_X}{\Sigma^{'}_Y}-{\Sigma^{'}_{XY}}^2\right)}\right).
% \end{equation} 
According to the standard equation of ellipse, the half length of an ellipse's \yr{major axis} is equal to the square root of the reciprocal of the \yr{smaller} eigenvalue ${\lambda}_{ad}$ of the ellipse's covariance matrix ${\Sigma}_{ad}$: $r_{ad}=\sqrt{1/{\lambda}_{ad}}=\sqrt{{2D}/(-\left(A+B\right)+\sqrt{\left(A-B\right)^2+C^2})}$. %by $\left|{\Sigma}_{ad}-{\lambda}_{ad} I\right|=0$. 
\rev{By substituting coefficients, }
% With elliptical coefficients substituted, 
adaptive radius is \yr{formulated} as follows:
\begin{equation}
    r_{ad}=\sqrt{2 {\lambda}^{'}_{max} ln\left(\frac{\sigma_i}{\alpha_{low}}\right)},
    \label{eq:r_ad}
\end{equation}
where ${\lambda}^{'}_{max}$ is the larger eigenvalue of the projected covariance ${\Sigma}^{'}$, which has been computed \wxz{earlier for the 99\%}
%during the original calculation of 
confidence interval (\yr{Detailed formula are presented in Supplementary materials}).

Finally, there may be situations where the Gaussian opacity is high or the predefined minimum splatting opacity is low, resulting in an adaptive radius $r_{ad}$ that exceeds the original radius $r_o$ determined based on the 99\% confidence interval. Therefore, the final value of the 2D Gaussian radius should be the minimum of these two values:
\begin{equation}
    r=min{\left(r_{ad},r_o\right)}.
\end{equation}

\yr{By performing early culling based on adaptive radius in \textit{Preprocess} stage in a parallel manner, we can save some computation costs caused by serial culling during \textit{Render} stage.}

\subsection{\yr{Early Culling with} Axis-aligned Bounding Box for Gaussian Splatting}
\label{ssec:aabb}

Although the adaptive radius significantly reduces unnecessary overhead in Gaussian splatting \yr{rasterization}, it \yr{cannot effectively cull} pairs of Gaussian-Tile with low splatting opacity in the minor axis direction \wxz{(Fig. \ref{fig:bboxs}b, regions marked in \yr{orange})}, as \yr{the radius} is determined by the major axis. 
To address this issue, we \yr{propose} axis-aligned bounding box for Gaussian splatting, which achieves a more significant reduction in ineffective expenses by accurately calculating the Gaussian size in the 2D directions.
% \begin{figure}[t]
%   \centering
%   \includegraphics[width=\linewidth]{figures/fig-aabb.pdf}
%   \caption{The rendering range of axis-aligned bounding box. For bounding circle , the adaptive radius determined by the semi-major axis length still retains unnecessary overhead in the minor axis direction, while the axis-aligned bounding box can solve this problem.}
%   \Description{To maintain rendering quality, the bounding circle algorithm adopts the semi-major axis length as the adaptive radius, resulting in pairs of Gaussian-Tile with overly low opacity along the short axis. The axis-aligned bounding box can reduce this overhead and alleviate the rendering workload.}
%   \label{fig:aabb}
% \end{figure}

For a projected Gaussian, \yr{we further cull the tiles based on}
the axis-aligned bounding box \yr{of the ellipse (Eq.~\eqref{eq:2d-ellipse})} %is also based on the 2D elliptic inequality 
constructed by the minimum splatting opacity ${\alpha}_{low}$ . 
Specifically, half of the bounding box's width $w$ and height $h$ \yr{equal to} the maximum values in the two coordinate directions of the ellipse, $x_{max}$ and $y_{max}$, \yr{respectively}. %To obtain them, we first rewrite the elliptic inequality into the corresponding function expression:
\yr{We first define a function $F$ for the ellipse:}
%\begin{equation}
    $F = A x^2+B y^2+C x y+D$.
%\end{equation}

Based on the \yr{ellipse} function, we calculate the corresponding partial derivatives in two coordinate directions respectively:
\begin{equation}
    \frac{\partial F}{\partial x} = 2 A x + C y, \ 
    \frac{\partial F}{\partial y} = 2 B y + C x.
\end{equation}

Based on geometric \yr{properties of ellipse}, when the ellipse function's partial derivative with respect to one coordinate direction is \wxz{0}, the coordinate value in another direction attains its extremum. 
To get the extremum of the ellipse on both coordinates, we \yr{let the two derivatives to be $0$, and} substitute the two coordinate relationships %corresponding to each derivative of zero 
into the ellipse function, and solve $x_{max}$ and $y_{max}$ as follows:
\begin{equation}
    {x}_{max}=\sqrt{2 {\Sigma}^{'}_X ln\left(\frac{\sigma_i}{\alpha_{low}}\right)}, \ 
% \end{equation}
% \begin{equation}
    {y}_{max}=\sqrt{2 {\Sigma}^{'}_Y ln\left(\frac{\sigma_i}{\alpha_{low}}\right)}.
\end{equation}

%To integrate with the Gaussian rasterization pipeline and cull pairs of Gaussian-Tile with low splatting opacity that persist in the bounding circle approach, 
\yr{Compared to bounding circle with adaptive radius,}
the axis-aligned bounding box \wxz{for Gaussian splatting} %enhances the 2D Gaussian radius to accommodate size in both the horizontal and vertical directions, 
\yr{can achieve different extents of culling in horizontal and vertical directions,}
thereby obtaining different tile ranges \yr{for the two directions}. Furthermore, similar to the adaptive radius for bounding circle, we take the original radius as the upper limit:
\begin{equation}
    r_x = min\left({x}_{max}, r_o \right),\ 
% \end{equation}
% \begin{equation}
    r_y = min\left({y}_{max}, r_o \right).
\end{equation}

With this approach, the Gaussian rendering range can align with the axis-aligned bounding box, significantly reducing rendering overhead and achieving more efficient rendering.

\subsection{Load Balancing for Pixel-parallel Splatting}
\label{ssec:load_balancing}

\begin{figure}[t]
  \centering
  \includegraphics[width=0.8\linewidth]{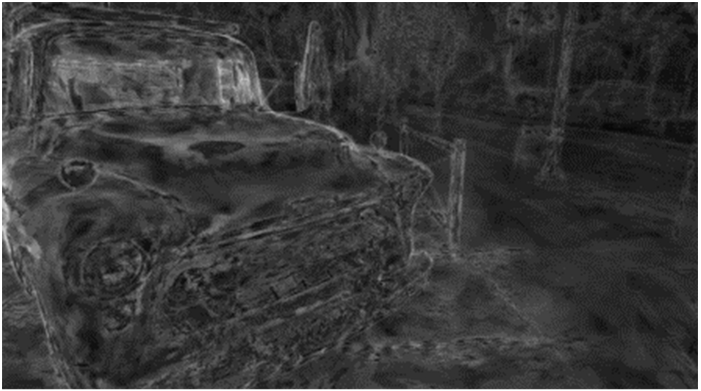}
  \caption{Visualization of pixel load in Truck dataset. Brighter pixels represent heavier load and darker pixels represent lighter load, which clearly shows the uneven load \yr{across pixel threads}.}
  \Description{A gray scale image with lots of bright and dark pixels, and the uneven distribution of pixels represent the uneven load.}
  \label{fig:pdistribution}
\end{figure}

\textit{Uneven Load Issue.} 
Based on the point-based rendering, %pixel coloring is characterized by its 
\yr{the \textit{Render} stage of Gaussian rasterization has a} pixel-parallel and Gaussian-serial architecture\yr{: the color calculation of each pixel is conducted in parallel, each pixel corresponding to a pixel thread; while in each pixel thread, the Gaussians are rendered sequentially}. 
Although \yr{the early cull by adaptive radius} and axis-aligned bounding box can \yr{accelerate \textit{Render} stage} by reducing the number of serial Gaussian, the rendering efficiency is still limited by the \textit{uneven load (computational time)} \yr{across} pixel threads \yr{(an example shown in Fig.~\ref{fig:pdistribution})}: 

% \begin{enumerate}
%     \item \yr{ Uneven load across tiles:} \yr{Since a 3D scene usually has different frequency information in different regions, where} \rev{regions with} {high}\rev{er} \yr{frequency requires more Gaussians,} %Due to the randomness in the distribution of high and low frequency information in 3D scenes, 
% the number of Gaussian \yr{of} each tile varies, leading to uneven load among \yr{different tiles.} %thread blocks as tiles correspond to thread blocks. 
%     \item \yr{ Uneven load within tiles:} Within %a GPU thread block, 
% \yr{a tile, for a certain Gaussian, different pixels have different splatting opacity due to different distances to Gaussian center. Since the front-to-back blending will be stopped when the accumulated opacity exceeds a threshold, the actual number of rendered Gaussian is different across pixels.}
% \end{enumerate}

\yr{1) Uneven load across tiles:} \yr{Since a 3D scene usually has different frequency information in different regions, where} \rev{regions with} {high}\rev{er} \yr{frequency requires more Gaussians,} %Due to the randomness in the distribution of high and low frequency information in 3D scenes, 
the number of Gaussian \yr{of} each tile varies, leading to uneven load among \yr{different tiles.} %thread blocks as tiles correspond to thread blocks. 
    
\yr{2) Uneven load within tiles:} Within %a GPU thread block, 
\yr{a tile, for a certain Gaussian, different pixels have different splatting opacity due to different distances to Gaussian center. Since the front-to-back blending will be stopped when the accumulated opacity exceeds a threshold, the actual number of rendered Gaussian is different across pixels.}
%different pixels will have different number of Gaussian to be rendered due to the different elimination of the same Gaussian. 
%Because pixels correspond to GPU threads, the load within thread block is also uneven. 

%An example of load visualization is shown in Fig.~\ref{fig:pdistribution}, where brighter pixels represent heavier loads and darker pixels represent lighter loads, indicating that the pixel thread load is significantly unbalanced. 
Based on the nature of GPU thread waiting, uneven load \yr{across pixel threads} lead\yr{s} to \yr{lower rendering} %running 
efficiency%, that is, slower rendering speed
~\cite{liu2015speculative}. 
%To improve rendering efficiency by enhancing the load balancing of 3D Gaussian rendering, 
\yr{To address the uneven load issue,} we \yr{propose} a \textit{load balancing} algorithm for Gaussian splatting.
\wxz{\yr{We first} reduce the number of Gaussian to be rendered for heavy-load pixels, \yr{and to address the rendering quality decrease caused by fewer Gaussians, we further} increase the number of Gaussian to be rendered for light-load pixels}.

%To begin with, it is crucial to grasp 
% \subsubsection{Pixel Loads Distribution.}
\textit{Pixel Loads Distribution.} 
\yr{We first compute}
the distribution of load \yr{of each pixel thread}, which is \yr{equivalent to the distribution of} the number of Gaussian ${\ell}_i$ to be rendered \yr{for} each pixel \wxz{:}%. %Utilizing the point-based rendering formula, 
%\yr{To render a single pixel from a viewpoint,}
%the number of Gaussian \yr{to be rendered} %needed to render a single pixel from a single viewpoint 
%can be 
% \yr{We formulate ${\ell}_i$} as follows:
\begin{equation}
     {\ell}_i=\sum_{j \in N}{\mathbbm{1}\left(G_j,p_i\right)},
\end{equation}
where %$G_j$ represents Gaussian $j$, $p_i$ represents pixel $i$, and 
\wxz{$N$ denotes the overall number of Gaussian and}
$\mathbbm{1}\left(G_j,p_i\right)$ \yr{is an indicator representing} whether Gaussian $j$ contributes to pixel $i$.

% \subsubsection{Load Balancing Loss.}
\textit{Load Balancing Loss.} 
Given that the \yr{rendering} efficiency %of the GPU 
under uneven loads hinges on heavy-load threads, \yr{to improve speed, the loads for these pixel threads need to be reduced.}
\yr{Meanwhile, encouraging} light-load threads to enhance their information \yr{by increasing Gaussians to be rendered} will not compromise efficiency. 
Thus, %for the 3D Gaussian Splatting model, 
\yr{our} load balancing \yr{strategy aims to} reduce the number of Gaussian to be rendered in heavy-load pixel\yr{s}, and increase that in light-load pixel\yr{s}. 
\yr{W}e adopt the standard deviation %as the estimation function for load balancing 
\yr{to estimate the difference of loads across pixels,}
\yr{and design the load balancing loss as follows}:
\begin{equation}
    {\mathcal{L}}_{load}=std_{i\in HW}\left({\ell}_i\right),
\end{equation}
where \wxz{H,W denotes the screen size and }
$std$ represents standard deviation function. The \yr{smaller $std$ indicates} more balanced load distribution, \yr{and} smaller difference in the number of Gaussian processed serially by each pixel thread, \yr{thus higher rendering efficiency}. %, and the smaller the value of ${\mathcal{L}}_{load}$. 
%Conversely, the larger the value of ${\mathcal{L}}_{load}$.

% \subsubsection{Total Loss.}
\textit{Total Loss.} 
To balance synthesis quality and rendering speed, we incorporate \yr{the load balancing loss} with L1-Loss and SSIM-Loss into the overall loss function, %thereby leveraging gradient back propagation to accelerate rendering and keep similar quality:
\yr{and the total loss is formulated as:}
\begin{equation}
    {\mathcal{L}}=\lambda_{L1}{\mathcal{L}}_1+\lambda_{ssim}{\mathcal{L}}_{ssim}+\lambda_{load}{\mathcal{L}}_{load},
\end{equation}
where $\lambda_{L1}$, $\lambda_{ssim}$, $\lambda_{load}$ are the weight coefficients of the L1 loss, SSIM loss, and load balancing loss, respectively. The numerical values satisfy $\lambda_{L1}+\lambda_{ssim}+\lambda_{load}=1$. 
% When $\lambda_{load}$ is large, the training system will pay more attention to load balancing and reduce the focus on rendering quality, resulting in an increase in rendering speed and a decrease in quality. On the contrary, the quality improves while the speed decreases.
% \rev{The optimization of ${\mathcal{L}}_{load}$ will propagate backwards to the geometric and opacity parameters of 3D Gaussians, indirectly affecting Gaussian densification and leading to a balanced load.}

\section{Experiments}
\begin{table*}
  \centering
  \caption{Quantitative evaluation of our method compared to the latest works, computed over the same full dataset as 3DGS. The bold results represent the best performance, while the underlined results stand for the second-best among all the results obtained in our own experiments. \textit{Ours-AABB} shows the results for our method with only the axis-aligned bounding box for Gaussian splatting.}
  \label{tab:Comparisons}
  
  \resizebox{0.95\textwidth}{!}{
  \begin{tabular}{c|cccccc|cccccc|cccccc}%
    % \toprule
    Dataset & \multicolumn{6}{c|}{Mip-NeRF360} & \multicolumn{6}{c|}{Tanks\&Temples} & \multicolumn{6}{c}{Deep Blending} \\
    Method|Metric & Train$\downarrow$ & FPS$\uparrow$ & Mem$\downarrow$ & SSIM$\uparrow$ & PSNR$\uparrow$ & LPIPS$\downarrow$ & Train$\downarrow$ & FPS$\uparrow$ & Mem$\downarrow$ & SSIM$\uparrow$ & PSNR$\uparrow$ & LPIPS$\downarrow$ & Train$\downarrow$ & FPS$\uparrow$ & Mem$\downarrow$ & SSIM$\uparrow$ & PSNR$\uparrow$ & LPIPS$\downarrow$  \\
    \midrule
     3DGS & 29m15s  & 185.85 & 780.7MB & \underline{0.813}  & \underline{27.49} & \underline{0.220} & 14m56s  & 253.39  & 437.5MB & \underline{0.844}  & \underline{23.64} & \textbf{0.178} & 24m14s  & 213.00  & 677.5MB & 0.899  & 29.50  & \textbf{0.246}  \\
     LightGaussian & 34m30s & 212.32 & \textbf{58.7MB} & 0.803 & 27.03  & 0.238 & 17m49s &  341.30 & \textbf{33.9MB} & 0.834  & 23.41 & 0.196 & 28m19s  & 259.15  & \textbf{52.0MB} & 0.896  & 29.12  & 0.258  \\
     C3DGS & 37m5s & 205.22 & {122.7MB} & 0.797 & 26.99 & 0.247 & 19m8s & 301.95 & {93.0MB} & 0.831 & 23.42 & 0.202  & 28m31s & 310.19 & {104.0MB} & \underline{0.902} & \underline{29.84} & 0.257\\
     EAGLES & 21m55s & 177.87 & \underline{68.11MB} & 0.805 & 27.14 & 0.239 & 11m22s & 300.68 & \underline{34.0MB} & 0.836  & 23.34 & 0.200 & 19m41s & 195.64 & \underline{61.5MB} & \textbf{0.907}  & \textbf{29.95}  & 0.249  \\
     GES & \underline{21m50s} & 241.67 & 332.9MB & 0.792  & 26.95  & 0.258 & \underline{10m56s} & 339.22  & 214.5MB & 0.836  & 23.36 & 0.197 & \underline{19m04s}  & 279.93  & 367.5MB & 0.901  & 29.60  & 0.252  \\
     \midrule
     Ours-AABB & 27m49s  & \underline{336.15}  & 786.7MB & \textbf{0.814}  & \textbf{27.51}  & \textbf{0.219} & 14m3s  & \underline{419.07}  & 445.5MB & \textbf{0.846}  & \textbf{23.76}  & \textbf{0.178} & 22m51s  & \underline{449.21}  & 693.0MB & 0.899  & 29.50  & \textbf{0.246}   \\
     Ours-Full & \textbf{17m44s}  & \textbf{589.61}  & 274.3MB & 0.783  & 26.90  & 0.272 & \textbf{9m2s}  & \textbf{718.42}  &  192.0MB & 0.832  & 23.53  & 0.205 & \textbf{15m32s}  & \textbf{716.46}  &  335.5MB & 0.901  & 29.65  & 0.254 \\
    % \bottomrule
  \end{tabular}
  }
\end{table*}

\begin{table*}
  \centering
  \caption{Ablation on same full dataset as comparisons. The experimentation begins with the original 3DGS project and sequentially introduces each sub method of ours. \rev{The bold results represent the best performance among all the results obtained in our ablation experiments.}}
  \label{tab:Ablation}
  
  \resizebox{1.0\textwidth}{!}{
  \begin{tabular}{c|cccccc|cccccc|cccccc}%
    % \toprule
    Dataset & \multicolumn{6}{c|}{Mip-NeRF360} & \multicolumn{6}{c|}{Tanks\&Temples} & \multicolumn{6}{c}{Deep Blending} \\
    Method|Metric & Train$\downarrow$ & FPS$\uparrow$ & Mem$\downarrow$ & SSIM$\uparrow$ & PSNR$\uparrow$ & LPIPS$\downarrow$ & Train$\downarrow$ & FPS$\uparrow$ & Mem$\downarrow$ & SSIM$\uparrow$ & PSNR$\uparrow$ & LPIPS$\downarrow$ & Train$\downarrow$ & FPS$\uparrow$ & Mem$\downarrow$ & SSIM$\uparrow$ & PSNR$\uparrow$ & LPIPS$\downarrow$  \\
    \midrule
     3DGS & 29m15s  & 185.85 & {780.7MB} & {0.813}  & 27.49 & {0.220} & 14m56s  & 253.39  & {437.5MB} & 0.844  & 23.64  & {0.178} & 24m14s  & 213.00  & {677.5MB} & 0.899  & 29.50  & \textbf{0.246}  \\
     3DGS\textbf{+}Adaptive radius & 28m38s & 221.90 & 782.2MB & {0.813}  & {27.50}  & {0.220} & 14m25s &  361.00 & 439.5MB & {0.845}  & {23.68} & \textbf{0.177} & 23m35s  & 320.10  & 678.5MB & 0.899  & 29.51  & \textbf{0.246}   \\
     3DGS\textbf{+}AABB & {27m49s}  & {336.15}  & 786.7MB & \textbf{0.814}  & \textbf{27.51}  & \textbf{0.219} & {14m3s}  & {419.07}  & 445.5MB & \textbf{0.846}  & \textbf{23.76}  & {0.178} & {22m51s}  & {449.21}  & 693.0MB & 0.899  & 29.50  & \textbf{0.246}   \\
     3DGS\textbf{+}AABB\textbf{+}Load balance (Ours) & \textbf{17m44s}  & \textbf{589.61}  & \textbf{274.3MB} & 0.783  & 26.90  & 0.272 & \textbf{9m2s}  & \textbf{718.42}  &  \textbf{192.0MB} & 0.832  & 23.53  & 0.205 & \textbf{15m32s}  & \textbf{716.46}  &  \textbf{335.5MB} & \textbf{0.901}  & \textbf{29.65}  & 0.254  \\
    % \bottomrule
  \end{tabular}
  }
\end{table*}

\subsection{Experimental Setting\yr{s}}
\subsubsection{Datasets and Metrics} 
To %clearly demonstrate the superiority 
\yr{validate the effectiveness} of our method, we utilize the same datasets and metrics as those used in 3DGS \cite{10.1145/3592433}. Specifically, the datasets comprises all scenes from Mip-NeRF360 \cite{9878829}, two scenes from Tanks\&Temples \cite{10.1145/3072959.3073599}, and two scenes from DeepBlending \cite{10.1145/3272127.3275084}, encompassing both bounded indoor scenes and unbounded outdoor environments. 
The \yr{evaluation} metrics include training duration, model size, rendering speed measured in frames per second (FPS), and synthesis quality assessed using PSNR, SSIM \cite{1284395}, and LPIPS \cite{8578166}. 
We employ identical hyperparameter settings across all experiments, with results reported on a single NVIDIA RTX 3090 GPU.

\subsubsection{Implementation}
Since adaptive radius and load acquisition in our method are based on Gaussian rasterization, these core methods are implemented using CUDA kernels, while load assessment and loss adjustments are handled by PyTorch. For bounding circle with adaptive radius, as the larger eigenvalue \yr{${\lambda}^{'}_{max}$} of the projected covariance ${\Sigma}^{'}$ is already computed, the only thing we need to do is \yr{to} multipl\yr{y} it by the opacity coefficient $2ln\left( \frac{{\sigma}_i}{{\alpha}_{low}} \right)$ instead of $3$ to yield the adaptive radius $r_{ad}$ \yr{as in Eq.~\eqref{eq:r_ad}}. %In terms of axis-aligned bounding box allowing for distinct tile ranges \yr{for the two directions}, the eigenvalues need to be replaced with the diagonal terms of the projected covariance. %It is important to 
\yr{N}ote that we maintain a minimum splatting opacity ${\alpha}_{low}$ of $\frac{1}{255}$ for optimal rendering quality, though higher values could further accelerate rendering. Regarding the load balancing approach, to prioritize quality optimization while accelerating rendering, the weight of its loss term ${\lambda}_{\rev{load}}$ is set to $0.45$, while the sum weight of remaining loss is reduced to $0.55$.

\subsection{Comparisons}

\wxz{%Besides the 3DGS baseline~ \cite{10.1145/3592433}, 
We compare \textit{AdR-Gaussian} (\textbf{Ours-Full} denotes \yr{our} full method\yr{,} and \textbf{Ours-AABB} denotes AABB-only method) to \yr{3DGS~\cite{10.1145/3592433}, and state-of-the-art 3DGS} faster rendering methods: LightGaussian \cite{fan2023lightgaussian}, C3DGS \cite{C3DGS}, EAGLES \cite{girish2023eagles}, \yr{and} GES \cite{GES}.}
% \yr{We show} 
\yr{The} qualitative \yr{and quantitative} comparisons \yr{are shown} in Fig. \ref{fig:comparisons} and Table \ref{tab:Comparisons}\yr{, respectively}. 

Despite varying amounts \rev{and distributions} of information across datasets, \wxz{our} \textit{AdR-Gaussian} \wxz{with only AABB for 3DGS} (denoted as \textbf{Ours-AABB}) achieves \wxz{higher} rendering \wxz{efficiency} \wxz{than any existing method} 
% and training speeds 
\wxz{on every dataset}. 
Specifically, %compared to 3DGS
% \wxz{for the FPS among test datasets}, 
early culling with AABB has significantly improved the rendering speed to 185\% of the \wxz{baseline \yr{(3DGS)}}, with an average rendering speed of \wxz{401} FPS \rev{on the three test datasets}. %310-675 %for \wxz{all} test scenes. 
% Training speed has also increased by an average of 37\%, and the model's average compression rate is 62\%. 
\wxz{Moreover, experiments demonstrate \textit{Ours-AABB} outperforms \yr{comparison} methods in \yr{rendering
% achieves equivalent or superior 
on Mip-NeRF360 and Tanks\&Temples datasets. Compared to the baseline 3DGS, we achieve lossless acceleration, which is due to} we only early cull Gaussian-Tile pairs that contribute nothing to the final color.}
%, while others' approaches may result in degeneration of quality.

\wxz{Additionally, our full \textit{AdR-Gaussian} method with both early culling and load balancing (denoted as \textbf{Ours-Full}) achieves the highest rendering speed among all methods, \yr{310\% on average compared to 3DGS.} 
This is realized by the reduction in the number of Gaussian to be rendered for heavy-load pixels, \textit{i.e.} reduction in thread waiting, %Due to the reduction in the number of Gaussian for heavy-load pixels, 
\yr{which also improves our} training speed and model size on every dataset compared to the baseline.
Meanwhile, \textit{Ours-Full} can achieve superior quality on \textit{Deep Blending} dataset, because \yr{our} load balancing also increases the number of Gaussian to be rendered for light-load pixels, which enhances the information on these pixels.} 

% the synthesis quality varies over dataset, with slight improvement on \textit{Deepblending} dataset and small decline on \textit{Mip-NeRF360} and \textit{Tanks\&Temples} datasets, where the improvement is achieved by enriching information of light-load pixels, and the decline is caused by the reduction in the number of Gaussian for heavy-load pixels. Due to the reduction in the number of Gaussian for heavy-load pixels, our training speed and model size are simultaneously improved for every dataset compared to the baseline. 
% while reducing the number of Gaussian for heavy-load pixels may lead to scene information loss during load balancing, we achieve \yr{equivalent rendering} quality by enriching information of light-load pixels.
% , while the method with only lossless early culling achieves the optimal quality.

% \wxz{Additionally, since \yr{the proposed} early culling approaches are efficient and lossless, we provide further comparisons between others and ours with only axis-aligned bounding box for Gaussian splatting, denoted as \textit{Ours-AABB}. While other methods cannot stably accelerate rendering across different dataset, ours with only early culling achieves faster rendering speed than existing methods as well as optimal rendering quality on every dataset.}

\begin{table}[t]
  \centering
  \caption{\rev{Experiments result on the same three dataset for the combination of \textit{Ours-Full} and the pruning startegy in LightGaussian.}}
  \label{tab:Combination}
  
  \resizebox{0.47\textwidth}{!}{
  \begin{tabular}{c|cc|cc|cc}%
    % \toprule
    \rev{Dataset} & \multicolumn{2}{c|}{\rev{Mip-NeRF360}} & \multicolumn{2}{c|}{\rev{Tanks\&Temples}} & \multicolumn{2}{c}{\rev{Deep Blending}} \\
    \rev{Method|Metric} & \rev{FPS$\uparrow$} & \rev{PSNR$\uparrow$} & \rev{FPS$\uparrow$} & \rev{PSNR$\uparrow$} & \rev{FPS$\uparrow$} & \rev{PSNR$\uparrow$} \\
    \midrule
     \rev{Ours-Full} & \rev{589.61} & \rev{26.90} & \rev{718.42} & \rev{23.53} & \rev{716.46} & \rev{29.65}  \\
     \rev{(+)Pruning} & \rev{741.12} & \rev{26.81} & \rev{1109.70} & \rev{23.42} & \rev{1035.37} & \rev{29.61} \\
    % \bottomrule
  \end{tabular}
  }
\end{table}

\subsection{Ablation Studies}
To clearly assess the performance of each component of our methods, \yr{we conduct} \wxz{ablation} \yr{studies as follows: we} begin with the original 3DGS project, and sequentially \yr{add} each \yr{module: early culling with adaptive radius, and with AABB for 3DGS, and load balancing}. 
The quantitative results are presented in Table \ref{tab:Ablation}, %.
\wxz{\yr{where} the contribution of each \yr{module} can be obtained by comparing corresponding rows.}

% \subsubsection{Adaptive Radius.}
\textit{Adaptive Radius.} 
The ablation experiment \wxz{shows} that 
\wxz{early-culling with adaptive radius enhances rendering speed for all test dataset\yr{s}, \yr{without rendering quality loss, and improving} training \yr{time}. 
Adaptive radius \yr{achieves rendering acceleration by moving} part of \yr{serial} culling from the Gaussian-serial \textit{Render} stage \yr{earlier} into the Gaussian-parallel \textit{Preprocess} stage. %, thus accelerating rendering speed. 
%On the other hand, such method 
\yr{Moreover, this method does not sacrifice rendering quality, achieving slight rendering quality improvement while largly improving speed.}
%is lossless and nothing to do with training, thus it should not have an impact on quality and other metrics.
}
% both methods can significantly accelerate the rendering speed of 3DGS models, with little impact on synthesis quality, training duration, and model size. This makes sense, as these methods only transfer the culling operation based on splatting opacity from the Gaussian-serial \textit{Render} stage to the Gaussian-parallel \textit{Preprocess} stage, which are lossless and efficient optimizations.

% \subsubsection{Axis-aligned Bounding Box.}
\wxz{\textit{Axis-aligned Bounding Box.}} 
\wxz{Experiment\yr{s} on three dataset\yr{s} demonstrate that \yr{early} culling with axis-aligned bounding box for Gaussian splatting achieves higher FPS, \textit{i.e.,} higher rendering efficiency. 
Since \yr{early-culling with AABB} can effectively cull the Gaussian-Tile pairs in the minor-axis direction, which \yr{can not be achieved by} the bounding circle calculated by adaptive radius, \yr{early-}culling with AABB moves more serial culling in \textit{Render} stage earlier into parallel culling in \textit{Preprocess} stage, leading to further acceleration.}
\rev{Experiments show that such early culling methods are lossless and suitable for any scene.}

% \subsubsection{Load Balancing.}
\textit{Load Balancing.} 
\wxz{Experimental result\yr{s} show load balancing method can \yr{further improve speed and} multiple metrics despite little sacrifice of quality.} \yr{Load balancing strategy} continuously reduces the the number of Gaussian to be rendered for heavy-load pixels with \yr{load balancing} loss, which enables a comprehensive optimization of rendering speed and \wxz{other metrics}. Additionally, load balancing also enriches information \yr{for} light-load pixels, which can offset the quality loss. 
% \rev{Experiments show that load balancing can achieve superior quality on \textit{Deep Blending} dataset.} 
% Depending on the information distribution, this algorithm can achieve equivalent or superior quality.
\rev{Regarding the application of load balancing in different scenes, it is more suitable for scenes with less high-frequency information, such as the scenes in the \textit{Deep Blending} dataset. The overall acceleration ratio increases with higher information frequency, sacrificing more quality, while local areas with relatively low-frequency information can always achieve more accurate modeling, hedging against the quality loss.}

\subsection{\rev{Discussions}}

\subsubsection{Limitation}
% While our Adr-Gaussian simultaneously optimizes rendering speed, training speed and model size, there are still some limitations. %to the method. 
% First, 
{The adjustment of minimum splatting opacity is not fully compatible with the training of 3D Gaussians.} 
Since Gaussian opacity is reinitialized every fixed intervals, setting an splatting opacity threshold \wxz{higher than this initialized value} can result in the culling of all Gaussians,
% whose initial opacities are smaller than the threshold, 
leading to rendering failures. Meanwhile, when only rendering is required, raising the splatting opacity threshold can safely speed up the process.

% \wxz{Moreover,} enriching information of light-load pixels may not fully compensate for the loss caused by the reduction in the number of Gaussian to be rendered in heavy-load pixels, resulting in quality degradation. \wxz{This could be addressed by adjusting the assessment of load balancing, which we leave as future work.}
% As load balancing is a well studied aspect of parallel computing, adjusting the assessment of load balancing may help maintain synthesis quality.
% \rev{Moreover, load balancing could blur those areas with high-frequency information and smooth color transitions, where larger Gaussians are used to replace the 
% numerous smaller Gaussians. As shown in \textbf{FIGXXX}, such ares could be closer to the GT value, but have}

% \subsubsection{\rev{Applicability in different scenes.}}
% \rev{Our method contains two phases: early culling and load balancing. Early culling is lossless and suitable for any scene. The effect of load balancing is influenced by the frequency of information: the overall acceleration ratio increases with higher information frequency, sacrificing more quality, while local areas with relatively low-frequency information can achieve more accurate modeling without significantly affecting rendering speed.}
% \rev{More specifically, for areas with high-frequency information and smooth color transitions, larger Gaussians are used to replace the numerous smaller Gaussians, which blurs these areas, shown in Fig.~\ref{fig:pipeline}.}

\subsubsection{\rev{Combination}}
% with Other Acceleration Methods
% \rev{Since Our \textit{AdR-Gaussian} mainly focuses on the rasterization pipeline of 3DGS, which is nothing to do with the Gaussian representation, thus it is easy to integrate our method with other 3DGS method, including those aimed at accelerating rendering. As shown in Table \ref{tab:Combination}, by adding the pruning strategy in LightGaussian, the rendering speed of \textit{Ours-Full} can be further accelerated with only a minor degeneration of quality.}
\rev{Our method focuses on the rasterization rather than the representation of 3D Gaussian, allowing for easy integration with other 3DGS methods, including those aimed at accelerating rendering. As shown in Table \ref{tab:Combination}, by adding the pruning strategy in LightGaussian, the rendering speed of \textit{Ours-Full} can be further accelerated with only a minor degeneration of quality.}
% the average FPS of \textit{Ours-Full} for the above three datasets increases from 628.94 to 843.09, with only a sacrifice of 0.08 PSNR.

\section{Conclusion}
\wxz{\yr{In this paper}, we propose \textit{AdR-Gaussian}, an acceleration for 3D Gaussian Splatting, which achieves a rendering speed of 310\% while maintaining equivalent or even better quality than its baseline (3DGS). By moving part of serial culling in \textit{Render} stage earlier into parallel culling in \textit{Preprocess} stage, and balances the load across different pixel threads to minimum thread waiting, \textit{AdR-Gaussian} is able to significantly accelerate the rendering speed of Gaussian \rev{S}platting. Compared with state-of-the-art fast rendering methods for \rev{3DGS}, our AdR-Gaussian shows much higher rendering efficiency.
% ours with only early-culling achieves higher rendering efficiency than all of them without rendering quality loss, while our full method proposes load balancing algorithm to minimum thread waiting, resulting in further acceleration.
}

\begin{acks}
We thank the reviewers for their constructive comments. This work was supported by National Natural Science Foundation of China (No. 72192821, 62302296, 62302297, 62272447), Shanghai Municipal Science and Technology Major Project (2021SHZDZX0102), Shanghai Sailing Program (22YF1420300), Young Elite Scientists Sponsorship Program by CAST (2022QNRC001), the Fundamental Research Funds for the Central Universities (project number: YG2023QNA35, YG2023QNB17, YG2024QNA44).
\end{acks}

\bibliographystyle{ACM-Reference-Format}
\bibliography{sample-base}
% \bibliography{sigconf-966}

\appendix

\begin{figure*}
  \centering
  \includegraphics[width=1.0\linewidth]{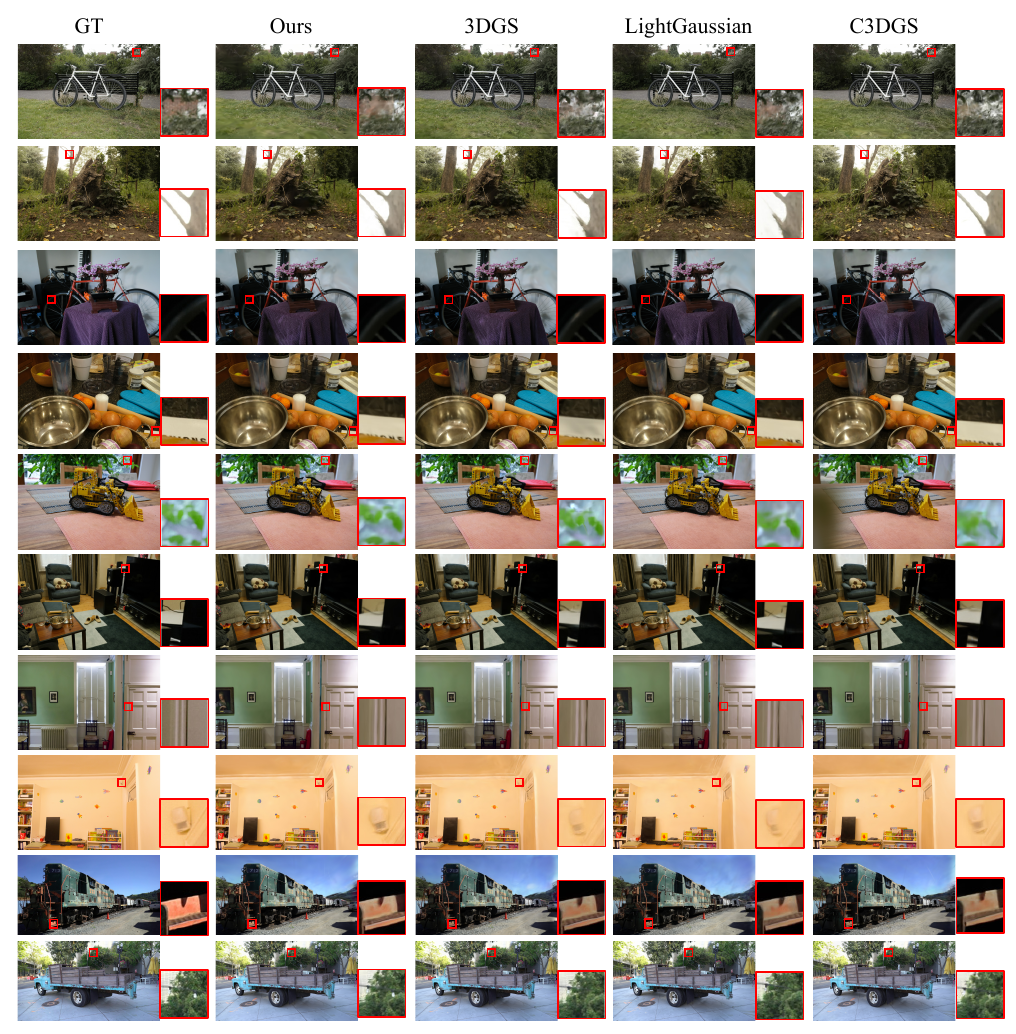}
  \caption{Our qualitative comparison with 3DGS and recent accelerated rendering methods, where images are from  held-out test views. The scenes are, from the top down: \textit{BICYCLE, STUMP, BONSAI, COUNTER, KITCHEN}, and \textit{ROOM} from the Mip-NeRF360 dataset; \textit{DRJOHNSON} and \textit{PLAYROOM} from the Deep Blending dataset; \textit{TRAIN} and \textit{TRUCK} from Tanks\&Temples dataset. \rev{Detailed comparisons are highlighted with red rectangles, where the larger rectangle is the zoomed-in view of the smaller one.} }
  \Description{In qualitative comparison, there is no significant degeneration of quality during the between the rendering results of our method and 3DGS and its related optimizations.}
  \label{fig:comparisons}
\end{figure*}

\end{document}